\documentclass{llncs}
\usepackage{llncsdoc}
\usepackage[pdftex]{graphicx}
\usepackage[cmex10]{amsmath}
\usepackage{algorithm}
\usepackage{algorithmic}
\usepackage{amssymb}
\usepackage{array}

\begin{document}

\title{Multi-Agent Optimization for Safety Analysis of Cyber-Physical Systems: Position Paper}

\author{\"{O}nder G\"{u}rcan, Nataliya Yakymets,\\Sara Tucci-Piergiovanni, Ansgar Radermacher}
\institute{CEA, LIST, Laboratory of Model driven engineering for embedded systems,\\
Point Courrier 174, Gif-sur-Yvette, F-91191 France\\
\email{\{onder.gurcan,nataliya.yakymets,sara.tucci,ansgar.radermacher\}@cea.fr}}

\maketitle

\begin{abstract}
Failure Mode, Effects and Criticality Analysis (FMECA) is one of the safety analysis methods recommended by most of the international standards. The classical FMECA is made in a form of a table filled in either manually or by using safety analysis tools. In both cases, the design engineers have to choose the trade-offs between safety and other development constraints. In the case of complex cyber-physical systems (CPS) with thousands of specified constraints, this may lead to severe problems and significantly impact the overall criticality of CPS. In this paper, we propose to adopt optimization techniques to automate the decision making process conducted after FMECA of CPS. We describe a multi-agent based optimization method which extends classical FMECA for offering optimal solutions in terms of criticality and development constraints of CPS. 
\end{abstract}

\section{Introduction}

Cyber-physical systems (CPS) are complex organizations of software and hardware systems (i.e. systems of systems) expected to serve, aid and cooperate with humans. 
Examples of CPS include large-scale engineering systems such as avionics, healthcare, transportation, automation and smart grids. 
CPS are expected to exceed traditional embedded systems in various aspects such as efficiency, safety, reliability, robustness, adaptability, availability and so on \cite{Park2012}. Since CPS are expected to interact and involve \textit{humans}, they require a high-level of safety which can be achieved by following rigorous procedures defined in safety standards \cite{IEC61508}. Those procedures describe application of various safety assessment (SA) methods starting from the early phases of system development life-cycle.

Over the last decade, the Model-Driven Engineering (MDE) \cite{Estefan2008} approach was widely used by design engineers to describe and analyze CPSs at the conceptual and design phases of their development cycle \cite{Selic2014}. Using open MDE frameworks, such as Eclipse Modeling Framework and Papyrus UML/SysML modeler, CPSs can be described in Unified Modeling Language (UML)\footnote{Object Management Group, The OMG Unified Modeling Language (OMG UML), Superstructure, version 2.4.1, 2011.}, System Modelling Language (SysML)\footnote{Object Management Group. OMG Systems Modeling Language (OMG SysML), 1st Sept. 2007.} or any Domain Specific Language (DSL) like AADL, RobotML, etc. and then extended to perform different types of analysis.  MDE environments integrate various technologies to provide an advanced support for system requirement management \cite{Adedjouma2010}, design \cite{Gerard2010}, analysis \cite{Yakymets2013,Walker2007,Walker2013}, verification and validation, and deployment. Most of those procedures, including SA activities,  can benefit from tighter coupling with MDE environments.

One of the typical SA methods recommended by safety standards \cite{IEC61508} is Failure Mode and Effects Analysis (FMEA) and its extensions: Production FMEA (or PFMEA), Criticality FMEA (or FMECA), Diagnostic FMEA (or FMEDA). PFMEA is used to prioritise, in terms of cost, problems to be addressed in system production. FMECA \cite{IEC60812} is used to assess the criticality of system  failures and to propose improvements introduced as a list of recommended preventive actions to avoid those failures. The criticality of system failure depends on failure severity, occurrence and detectability and will be defined in the next section. By the preventive action we mean a change in system architecture that may be implemented to address a failure of system components. Another modification of FMEA called FMEDA helps safety experts to define self checking features and provides detailed recommendations for system architecture. Therefore, FMEA is used to identify causes and effects of failures that might appear across system life-cycle and, in addition, it gives a detailed specification of system failures, their criticality, cost and ways to avoid them. However, in the case of complex CPS with thousands of specified failures, multiple FMEA tables (PFMEA, FMECA, FMEDA) may lead to severe problems when choosing the trade-offs between cost, criticality and diagnostic coverage of a system. Therefore, existing methods and tools for FMEA can be improved to automate this task and to offer optimal solutions in terms of safety related constraints.

In this paper we focus on FMECA type of analysis. The classical FMECA is made in a form of the table filled in either manually or by using MDE methods and tools for SA \cite{Yakymets2013,Walker2007}. The latter approach helps to partially automate quantitative part of FMECA related to \textit{criticality} assessment of system components. However, FMECA is not capable to detect cases when i) multiple failures can be rectified using the same  preventive action or ii) several alternative prevented actions have been defined to rectify a single failure. 
This may lead to overuse of resources needed to increase system safety (implement redundant preventive actions) and can significantly impact the cost of CPS. The design engineers have to decide which actions might be rejected to reach a trade-off between \textit{safety} and other development constraints. Therefore, existing methods and tools for FMECA can be improved to automate this task and to offer optimal solutions in terms of \textit{criticality} and other constraints.  
In the literature, this sort of optimization is called multidisciplinary optimization (MDO). The goal of MDO is to find the configuration that maximizes (or minimizes) several objectives while satisfying several constraints \cite{Sobieszczanski-Sobieski1997}.

We, therefore, propose to extend the classical FMECA using MDO to resolve the aforementioned problem. The proposed method analyzes recommended preventive actions associated with component failures and categorizes them according to existing constraints. As an MDO approach, we have chosen the Adaptive Multi-Agent Systems (AMAS) approach \cite{Capera2003} since it is a good candidate for finding an optimal level of system \textit{criticality} and \textit{configuration} \cite{Kaddoum2012}. Using AMAS, we can find a set of configurations where the \textit{criticality} of each component is below the threshold as much as possible under certain development constraints. We believe that the use of a criticality-based self-adaptation technique like AMAS, along with the adoption of SA knowledge will make it possible to harness the complexity of the given problem by finding \textit{optimal} configurations automatically. 

The remaining of the paper is organized as follows. Section 2 gives background information about FMECA and states the problem. Section 3 presents the AMAS approach and shows our method to extend classical FMECA to build safety self-adaptable CPSs. Lastly, Section 4 discusses and concludes the paper and gives some prospects for further work.

\section{\label{sec:Problem}Failure Mode, Effects and Criticality Analysis}

FMECA is an inductive bottom up approach used to identify different effects (or consequences) and causes of component failures by analyzing them from component-level up till system-level. While the qualitative FMEA \cite{IEC60812} helps to define main causes and effects of failures, the quantitative FMECA identifies the \textit{criticality} level of failures. The \textit{criticality} of a failure is automatically evaluated according to (1). 

\begin{eqnarray}
\mathcal{C}(f) = \mathcal{S}(f) \cdot \mathcal{O}(f) \cdot \mathcal{D}(f),
\end{eqnarray}

where $\mathcal{C}$ is the criticality, $\mathcal{S}$ is the severity, $\mathcal{O}$ is the occurence and $\mathcal{D}$ is the detectability of failure $f$. The severity of failure characterizes the consequences that the failure could have on CPS or its environment. The occurence of failure characterizes the appearance of the failure and its average exposure occurrence on CPS or its environment. The detectability of the failure characterizes the means which exist to detect or plan the appearance of the feared event. The severity, occurence and detectability criteria are usually evaluated according to the matrices recommended by the domain specific standards and norms. Table \ref{tab:detectability} gives the evaluation matrices for the severity, occurence and detectability criteria  adopted in our work. 

\begin{figure*}[!t]
\centering
\includegraphics[scale=0.87]{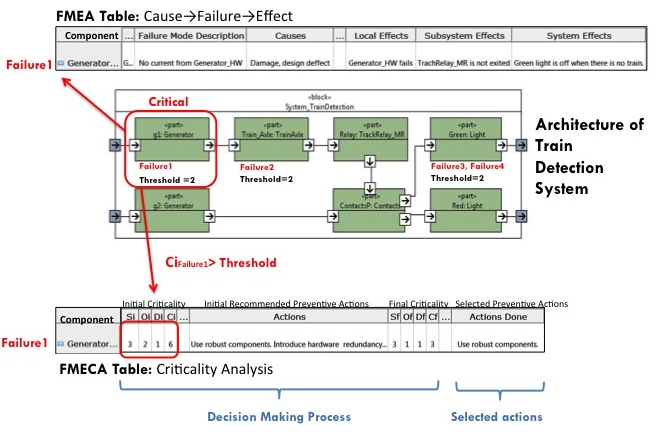}
\caption{A simple example of FMEA and FMECA.}
\label{fig:FMEA-FMECA}
\end{figure*}

During FMECA each system component is annotated with the \textit{critical threshold}, the boundary value of the allowed \textit{criticality} for this component. If the value of \textit{criticality} estimated through FMECA is higher than the critical threshold, the analyzed component is considered as \textit{critical}. Figure \ref{fig:FMEA-FMECA} illustrates a simple example of FMEA and FMECA: The FMEA table describes possible causes and effects of the failure called Failure1 ("No current from Generator") of the Generator component of the train detection system. The FMECA table shows the results of the criticality analysis of Failure1 in the Generator component. The evaluation matrices scale severity, occurence and detectability criteria from 1 to 4 (Table \ref{tab:detectability}).  We assume that the critical threshold of each component is 2. According to (1) the estimated initial criticality of Failure1 is 6 which is higher than the threshold. Consequently, Failure1 is critical for the Generator component. The two actions recommended to tackle Failure1 of  Generator include "Use of robust components" and "Introduction of hardware redundancy".

\begin{table*}
\caption{\label{tab:detectability}Evaluation matrices for severity, occurence and detectability.}
\begin{centering}
\begin{tabular}{|c|>{\centering}m{2.5cm}|>{\centering}m{2.5cm}|>{\centering}m{2.5cm}|>{\centering}m{2.5cm}|}
\hline 
\multicolumn{5}{|c}{\textbf{Severity}}\tabularnewline
\hline 
\textbf{Level} & Negligible & Significant & Critical & Catastrophic\tabularnewline
\hline 
\textbf{Rank} & 1 & 2 & 3 & 4\tabularnewline
\hline 
\textbf{Descr.} & Deterioration of the system with no impact on its availability neither functioning. & Deterioration of the system, which makes it not available to perform some operations. & Deterioration of the system, which leads to its unavailability, permanent or definitive. & User's deadly, potentially deadly or permanent injuries.\tabularnewline
\hline 
\hline 
\hline 
\multicolumn{5}{|c}{\textbf{Occurence}}\tabularnewline
\hline 
\textbf{Level} & Very Low & Low & Medium & High\tabularnewline
\hline 
\textbf{Rank} & 1 & 2 & 3 & 4\tabularnewline
\hline 
\textbf{Descr.} & Less than once a week. & At least once a week. & Several times a week. & Daily.\tabularnewline
\hline 
\hline 
\hline 
\multicolumn{5}{|c}{\textbf{Detectability}}\tabularnewline
\hline 
\textbf{Level} & High & Medium & Low & Very Low\tabularnewline
\hline 
\textbf{Rank} & 1 & 2 & 3 & 4\tabularnewline
\hline 
\textbf{Descr.} & Failure mode systematically detectable before its appearance. & Failure mode usually detectable before its appearance. & Failure mode hardly detectable before its appearance. & Failure mode not detectable before its appearance.\tabularnewline
\hline 
\end{tabular}
\par\end{centering}
\end{table*}

\begin{figure}[!t]
\centering
\includegraphics[scale=0.83]{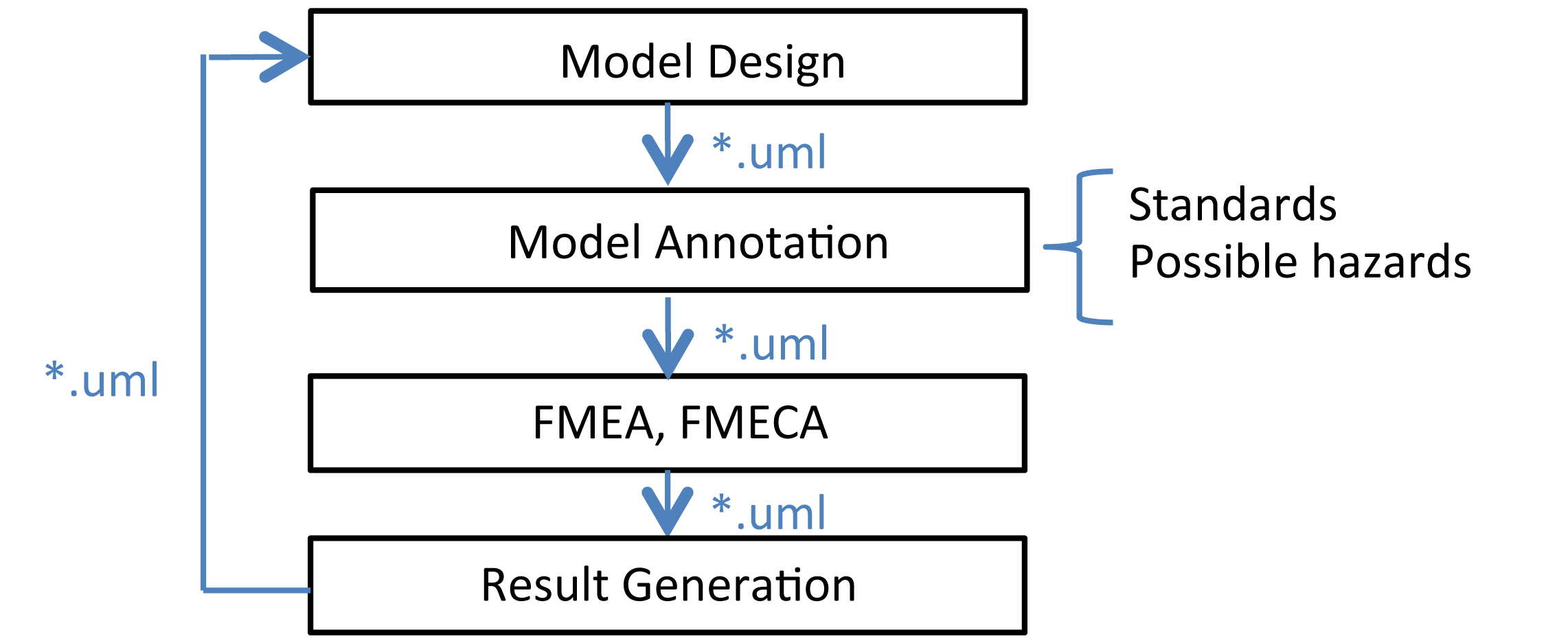}
\caption{The automated FMEA process.}
\label{fig:Automated-FMEA}
\end{figure}

The FMECA process offered by most of the MDE tools for SA can be summarized in several steps: system modeling, annotation, analysis, and result generation (Figure \ref{fig:Automated-FMEA}). A system model is created using either UML/SysML notations (e.g., Sophia \cite{Yakymets2013} or HiP-HOPS \cite{Walker2007} tools) or formal languages like NuSMV \cite{Bozzano2007}, SAML \cite{Gudemann2010}. Then the model is annotated (or extended) with the description of possible failures of system components for further FMECA which is conducted according to the specified failures and their severity, occurrence and detectability criteria. The results are displayed using dedicated profiles, editors, tables and report generation modules.

The classical FMECA gives a detailed description of causes and effects of single failures on component-level, checks what happens on system-level and guides design and safety engineers how to prevent failures. The latter is done via analysis and selection of recommended \textit{preventive actions} associated with every failure. A preventive action is a change in system architecture that may be \textit{selected} and further \textit{implemented} to address a failure of system components. We consider several types of relations between recommended actions and failure modes: 
\begin{itemize}
\item Relation type 1: one action targets one failure mode; 
\item Relation type 2: several complementary actions target one failure mode;
\item Relation type 3: several alternative actions target the same failure mode;
\item Relation type 4: one action targets several failure modes. 
\end{itemize}

Those actions are defined during the FMECA process by the safety engineer and can be implemented by the design engineer to improve system safety. 

However, in the case of strict specification requirements, such as \textit{cost} (how much it will cost to implement all the preventive actions) or \textit{time} constraints (how long it will take to implement all the preventive actions), some improvements recommended during FMECA may be rejected (for instance, when choosing between several alternative actions targeting the same failure). Taking into account a complexity of CPSs and, as a result, sophisticated FMECA (such FMECA tables can include thousands of failures and associated preventive actions) the choice of preventive actions that should be implemented is not trivial. In addition, this can significantly increase the overall \textit{criticality} of CPS. The FMECA tools do not provide information on how to optimize a set of preventive actions that should be selected for further implementation. 

Based on this observation, we propose to extend classical FMECA method to automate the \textit{decision making process} related to choosing optimal \textit{selected actions} in terms of safety and other development constraints (Figure \ref{fig:FMEA-FMECA}). We describe a method for optimized safety analysis based on FMECA technique and show our preliminary model that is intended to be implemented by extending our tool for SA called Sophia \cite{Yakymets2013}.

\section{Using AMAS for Improving FMECA}

For improving the FMECA process, an optimization technique, that will help finding an optimum set of recommended actions by taking into account the trade-off between criticality and cost, is needed. 
In this sense, the Adaptive Multi-Agents Systems (AMAS) approach \cite{Capera2003} has been selected since it fits well to the complexity of the problem at hand \cite{Jorquera2013}. 

\subsection{The AMAS Approach}

\subsubsection{Overview}

In the AMAS approach, the system is composed of a set of dynamic number of autonomous agents $\mathcal{A}=\{a_0, a_1, ...\}$. 
According to AMAS, a system is said to be \textit{functionally} adequate if it produces the function 
for which it was designed, according to the viewpoint of an external observer who knows its finality. 
To reach this functional adequacy, it has been proven that each autonomous agent $a_i \in \mathcal{A}$ 
must keep relations as cooperative as possible with its social (other agents) and physical environment \cite{Capera2003,Camps1998}.
To do so, each agent $a_i$ keeps tracks of its degree of criticality and tries to help the most \textit{critical} agent in its neighborhood including itself. 
In other words, a critical agent is said to be the most \textit{dissatisfied} one and its criticality has to be reduced either by itself or its neighbors\footnote{In certain conditions, it spontaneously communicates information to agents that it thinks the information would be useful.}.

\subsubsection{Agent-Criticality Heuristic}

The \textit{criticality}\footnote{To avoid confusion of the term \textit{criticality} between the safety analysis domain and the AMAS approach, the terms \textit{safety-criticality} and \textit{agent-criticality} are used respectively hereafter in this paper.} value of an agent $a_i$ at time $t$ is calculated by using an \textit{agent-criticality} function $c_{a_i}(t)$. This function may return a value ranging between $0.0$ and $100.0$ where $0.0$ criticality indicates the highest degree of satisfaction and $100.0$ criticality indicates the lowest degree of satisfaction of an agent. 
There is no single formula for \textit{agent-criticality} functions.
It can be defined by making use of the agent's internal parameters and state: e.g., an agent with correct internal parameter values is said to be closer to its goal and thus should be more satisfied than another agent that is still searching for its right parameter values; or an agent in a non-cooperative state should be less satisfied compared to another agent in a cooperative state.
However, evaluation methods and calculation of the \textit{agent-criticality} are specific to each type of cooperative agent and thus may change from domain to domain.
Consequently, it is the designer's responsibility to identify the most appropriate \textit{agent-criticality} function for each type of cooperative agent
depending on the problem, the domain, the constraints, etc.

\subsubsection{Non-Cooperative Situations}

The value of inputs coming from other agents (and physical environment) leads $a_i$ to produce a new decision. 
A non-desired configuration of inputs causes a non-cooperative situation (NCS) to occur.
$a_i$ is able to memorize, forget and spontaneously send feedbacks 
related to desired or non-desired configurations of inputs coming from other agents. 
We denote the set of feedbacks as $\mathcal{F}$ and model sending a feedback $f_a \in \mathcal{F}$ 
using the action of the form {\small \textsf{send($f_a, \mathcal{R}$)}} 
where $a$ is the source of $f$ and receiver agents $\mathcal{R} \subset \mathcal{A}$ \textbackslash{} $\{a\}$. 
A feedback $f_a \in \mathcal{F}$ can be about increasing the value of the input ($f_a${\ensuremath{\uparrow}}),
decreasing the value of the input ($f_a${\ensuremath{\downarrow}})
or informing that the input is good ($f_a${\ensuremath{\approx}}).

\subsubsection{Local Solving}

When a feedback about a NCS is received by an agent, at any time during its life-cycle, 
it acts in order to avoid or overcome this situation \cite{Bernon2009} for coming back to a cooperative state. 
This provides an agent with learning capabilities and makes it constantly adapt to new situations that are judged harmful. 
In case a NCS cannot be overcome by an agent, it keeps track of this situation by using a level of annoyance value 
$\psi_{f_a}$ where $f_a$ is the feedback about this NCS.
When a NCS is overcome, $\psi_{f_a}$ is set to $0$, otherwise it is increased by $1$.
The first behaviour an agent tries to adopt to overcome a NCS is a \textit{tuning behaviour} in which it tries to
adjust its internal parameters.
If this tuning is impossible (because a limit is reached or the agent knows that a worst situation will
occur if it adjusts in a given way), it may propagate the feedback (or an interpretation of it) to other agents that may handle it. 
If such a behaviour of tuning fails many times and $\psi_{f_a}$ 
crosses the reorganization annoyance threshold $\psi_{reorganization}$ (reorganization condition), 
an agent adopts a \textit{reorganisation behaviour} in which it tries to change the way of its interaction
with others (e.g., by changing a link with another agent, by creating a new one, by changing the way in which it communicates with another
one, etc.) In the same way this behaviour may fail counteracting the NCS and a last kind of behaviour may be adopted
by the agent: \textit{evolution behaviour}. 
This is detected when $\psi_{f_a}$ crosses the evolution annoyance threshold $\psi_{evolution}$ (evolution condition).
In the evolution step, an agent may create a new one (e.g., for helping itself because it found nobody else) 
or may accept to disappear (e.g., it was totally useless and decides to leave the system). 
In these two last levels, propagation of a problem to other agents is always possible if a local processing is not achieved. 
The overall algorithm for suppressing a NCS by an agent is given in  Algorithm 1 in \cite{Gurcan2014}.

\subsection{Identification of agents and their nominal behaviors}

We designed an agent-based simulation model $Sim$, for optimizing FMECA described in Section \ref{sec:Problem}, by basically capturing all taken design decisions based on the AMAS theory as a dynamic undirected graph $Sim(t)=(\mathcal{G}(t), \mathcal{P}(t),q)$ where $\mathcal{G}(t)$ is the set of time varying failure mode agents, $\mathcal{P}(t)$ is the set of time varying preventive action agents and $q$ is the quality agent.

In the initial model, we only consider the \textit{cost} constraints. Consequently, each preventive action agent $p \in \mathcal{P}(t)$ has only a cost parameter. If $p$ is selected the cost has a non-zero value, otherwise its value is zero. 

Each failure mode agent $g \in \mathcal{G}(t)$ has a set of recommended preventive action agents $\mathcal{P}_g(t) \subset \mathcal{P}(t)$ and selects preventive action agents $p \in \mathcal{P}_g(t)$ for implementation. 
In the initial model, we considered the cases where \textit{one action can target one failure mode} (Section 2, relation type 1) and \textit{several complementary actions target one failure mode} (Section 2, relation type 2).
We denote the set of preventive actions of a failure mode agent $g$ at time $t$ as $Sel_g(t) = \{p \in \mathcal{P}_g(t) | \{p_1, p_2, .., p_n\}\}$. On the other hand, each preventive action $p \in \mathcal{P}_g(t)$ is selected by failure mode agents $g \in \mathcal{G}(t)$. We denote the set of failure mode agents of a preventive action agent $p$ at time $t$ as $SelBy_p(t) = \{g \in \mathcal{G}(t) | \{g_1, g_2, .., g_n\}\}$. 

A failure mode agent $g \in \mathcal{G}(t)$ through its nominal behaviour aims to increase the number of selected preventive actions, as much as possible. Similarly, a preventive action agent $p \in \mathcal{P}(t)$ through its nominal behaviour aims to increase as much as possible the number of failure modes it is selected for.

The quality agent $q$ is responsible for the satisfaction of the global quality properties like \textit{cost} and \textit{time} constraints. $q$ knows list  all preventive action agents $p \in \mathcal{P}(t)$. Since we only consider \textit{cost} in this paper, the nominal behaviour of $q$ is to continuously collect the cost information from the preventive action agents $p \in \mathcal{P}(t)$, calculate the total cost $\tau(t)$ for each time $t$ by summing up the cost of each $p \in \mathcal{P}(t)$ and compare it with the total project budget $\beta$. 

%
%
%

\subsection{Identification of Non-Cooperative Situations and Feedbacks}

The proposed agent-based model, in which the configuration of failure mode agents and preventive action agents (their number and connection) can change, is subject to NCSs.
All NCSs are identified by analyzing the possible bad situations of FMECA regarding to the explanation given in Section \ref{sec:Problem}.

\subsubsection{\label{sec:BadCriticalityValue-NCS}Bad Safety-Criticality Value}

If the number of selected preventive actions of a failure mode agent $g$ is not enough and thus the safety-critical threshold is crossed, a \textit{bad safety-criticality value NCS} is detected by $g$. 
When such a situation is detected at time $t$, the failure mode agent $g$ should improve its preventive actions set (by having better preventive actions in the set). 
To do so, it sends an \textit{select more} feedback ($f \in \mathcal{F}_{sel}${\ensuremath{\uparrow}}) to some or all of its recommended preventive action agents $p \in P_g(t)$. 
Otherwise, the selection is good and an \textit{selection good} feedback 
($f \in \mathcal{F}_{sel}${\small {\ensuremath{\approx}}}) is sent to $P_g(t)$.

\subsubsection{\label{sec:BadCost-NCS}Bad Total Cost}

The quality agent $q$ continuously calculates the total cost $\tau(t)$ as mentioned before and if this cost crosses the total budget $\beta$ ($\tau(t)>\beta$) a \textit{bad total cost NCS} is detected. 
When such a situation is detected at time $t$, the quality agent $q$ should reduce the number of selected preventive actions (and, consequently, the budget). 
To do so, it sends a \textit{select less} feedback ($f \in \mathcal{F}_{sel}${\ensuremath{\downarrow}}) to some or all of its preventive action agents $Pre_q(t)$. 
Otherwise, the costs are good and a \textit{selection good} feedback 
($f \in \mathcal{F}_{sel}${\small {\ensuremath{\approx}}}) is sent to $Pre_q(t)$.

\subsection{Agent-Criticality Functions}

As described before, each cooperative agent $a$ has to define an \textit{agent-criticality} function $c_{a}(t)$ for calculating their \textit{agent-criticality} value at time $t$.

\subsubsection{Failure Mode Agents}
For a failure mode agent $g_i$ the agent-criticality is inversely proportional to the number of its selected preventive actions, thus as an agent-criticality function we use

\begin{equation}
	c_{g_i}(t)=\frac{1}{m_i(t)}
\end{equation}

where $m_i$ is the number of selected preventive actions of $g_i$ at time $t$.

\subsubsection{Preventive Action Agents}
Similarly, for a preventive action agent $p_i$ the agent-criticality is inversely proportional to the number of failure modes it is selected for, thus as an agent-criticality function we use

\begin{equation}
	c_{p_i}(t)=\frac{1}{n_i(t)}
\end{equation}

where $n_i$ is the number of failure modes of $p_i$ at time $t$.

\subsubsection{Quality Agent}
For the quality agent $q$, on the other hand, the agent-criticality is inversely proportional to the number of its selected preventive actions, thus as an agent-criticality function we use

\begin{equation}
	c_{q}(t)=\frac{1}{n_i(t)}
\end{equation}

where $n_i$ is the number of failure modes of $p_i$ at time $t$.


\subsection{Cooperative Behaviours}

There is no \textit{tuning behaviour} for agents in our initial model since there is no parameter to tune. 
There is also no \textit{evolution behaviour} for any agent type since the failure modes and recommended preventive actions are predefined. Currenty, we only defined \textit{reorganization behaviours} as cooperative behaviours of agents.

The \textit{reorganization behaviours} of failure mode agents and recommended preventive action agents are modelled using actions of the form {\small \textsf{add(${\{g, p\}}$)}} and {\small \textsf{remove($\{g, p\}$)}} for $g \in{} \mathcal{G}(t)$ and $p \in{} \mathcal{P}_{g}(t)$,
which correspond to the formation and suppression (respectively) of a selection relation $\{g, p\}$ at time $t$. 
It is assumed that no \textit{selection relation} is both added and removed at the same time.

NCSs are suppressed by processing these cooperative behaviours as follows.
When a select more feedback ($f \in \mathcal{F}_{sel}${\ensuremath{\uparrow}}) is received by a preventive action agent $p \in{} \mathcal{P}_{g}(t)$ from a failure mode agent $g$, $p$ first checks if it is the most critical preventive action agent among its neighbours. If yes, it executes {\small \textsf{add(${\{g, p\}}$)}} both for helping $g$ and reducing its criticality. If there are more critical neighbours, $p$ forwards the incoming feedback to the most critical one.

Similarly, when a select less feedback ($f \in \mathcal{F}_{sel}${\ensuremath{\downarrow}}) is received by a preventive action agent $p \in{} \mathcal{P}_{g}(t)$ from a failure mode agent $g$, $p$ first checks if it is the most critical preventive action agent among its neighbours. If yes, $p$ forwards the incoming feedback to the most critical one. If there are more critical neighbours, it executes {\small \textsf{remove(${\{g, p\}}$)}} both for helping $g$ and its neighbours.

Otherwise, if a \textit{selection good} feedback ($f \in \mathcal{F}_{sel}${\small {\ensuremath{\approx}}}) is received, $p$ does not execute any cooperative behaviour.

\section{Discussion \& Conclusions}

Cyber-physical systems (CPS) are constantly growing in complexity. This is accompanied by an increasing need for safety in those systems. In this paper, we propose to enrich the safety assessment process for CPSs by improving the Failure Mode, Effects and Criticality Analysis (FMECA) method. Our motivation is based on the observation that the classical FMECA becomes very complicated and time consuming when the system at hand has thousands of failure modes and thousands of corresponding recommended preventive actions. In addition to the classical FMECA, the method proposed in this paper adopts a multi-agent approach, namely Adaptive Multi-Agent Systems (AMAS), for providing an optimal set of recommended actions to be implemented, in order to get a trade-off between system \textit{safety-criticality} and other constraints such as \textit{cost} and \textit{time}. 

AMAS is a promising candidate for \textit{criticality}-based optimization problems like one arising from post analysis of FMECA results. 
Moreover, since AMAS is a self-organizing solution, it offers significant advantages such as increased scalability \cite{Martin-Flatin2006}. It distributes the complexity of the preventive action selection issue across agents. The scalability and success on optimization of AMAS has been shown before in various studies \cite{Couellan2015,Gurcan2014,Jorquera2013,Gurcan2012,Kaddoum2012,Combettes2012,Welcomme2009}. 
Due to the limitations coming from the current state of the art of the FMECA methods and tools, it is not possible to realize the proposed solution directly. First, we need information about constraints such as cost, time etc. which is not the scope of classical FMECA. Secondly, we need to be able to automate the calculation of \textit{safety-criticality} per failure mode depending on the implemented actions. 

In the initial model presented in this paper, we only considered the cases where \textit{one action can target one failure mode} (Section 2, relation type 1), \textit{several complementary actions target one failure mode} (Section 2, relation type 2) and the \textit{cost} constraints. 
We will further elaborate our model for covering all the relation types between recommended preventive actions and failure modes, and the \textit{timing} constraints.
This way, we plan to obtain a highly realistic model.

Furthermore, as a future prospect, we believe that if we can integrate more safety related information (such as behavioral model) to the safety analysis process, it would be possible to realize fully-automated solutions for FMECA with high coverage of situations. For instance, such a solution may take into account \textit{run-time} resource consumptions while deciding the number and type of replications needed. Currently, there is no such model in the literature.

\bibliographystyle{splncs}
\bibliography{references}
%
\end{document}